\documentclass[preprint,12pt]{elsevier}

\usepackage[utf8]{inputenc} 
\usepackage[T1]{fontenc}    
\usepackage{hyperref}       
\usepackage{url}            
\usepackage{booktabs}       
\usepackage{amsfonts} 
\usepackage{graphicx}
\usepackage{nicefrac}       
\usepackage{microtype}      
\usepackage{xcolor}         

\title{NLP Inspired Training Mechanics For Modeling Transient Dynamics}

\author[anss]{Lalit Ghule}

\affiliation[anss]{organization={Office of CTO},
            addressline={Ansys Inc.}, 
            city={Canonsburg},
            postcode={15317}, 
            state={PA},
            country={USA}}

\affiliation[anss2]{organization={Office of CTO},
            addressline={Ansys Inc.}, 
            city={San Jose},
            postcode={95134}, 
            state={CA},
            country={USA}}

\author[anss]{Rishikesh Ranade}
\author[anss2]{Jay Pathak}

\begin{document}

\maketitle

\begin{abstract}
  In recent years, Machine learning (ML) techniques developed for Natural Language Processing (NLP) have permeated into developing better computer vision algorithms. In this work, we use such NLP-inspired techniques to improve the accuracy, robustness and generalizability of ML models for simulating transient dynamics. We introduce teacher forcing and curriculum learning based training mechanics to model vortical flows and show an enhancement in accuracy for ML models, such as FNO and UNet by more than 50\%.  
\end{abstract}

\section{Introduction}
Numerical simulations are proving to be of paramount importance across different avenues of industrial and research development. These simulations are performed by solving partial differential equations, which are represented on a discretized computational domain using finite difference or finite volume methods. These methods provide accurate predictions, but they are computationally very expensive. As a result, researchers in the deep learning community have devised many different models to learn physics behind these engineering problems using supervised learning methods, that determine the input to output mapping \cite{li2020fourier, bhatnagar2019prediction, zhu2018bayesian, guo2016convolutional, ranade2022composable} or unsupervised learning methods, that embed physical laws into loss functions to compute PDE solutions \cite{bar2019unsupervised, smith2020eikonet, raissi2019physics, ranade2021discretizationnet}. These physics-informed methods provide a unique benefit over most approaches by imposing initial and boundary conditions in the optimization process. Even though these surrogate models perform reasonably well, they suffer from error accumulation, especially in the extrapolation regime. The accumulation of error is worse for transient problems because the deep learning models diverge during inference. Many researchers have tried to solve this problem by proposing different operator networks like Fourier Neural Operator (FNO) \cite{li2020fourier}, Multiscale Neural operator \cite{lutjens2022multiscale}, Koopman operator \cite{balakrishnan2021stochastic} etc. but the error accumulation problem for long time range predictions is still prevalent. 

This problem is very common in the field of natural language processing. In the models that translate one language to another, error-prone predictions in the beginning lead to a completely different output during sequential rollouts. To tackle this problem, researchers in this field have come up with several approaches. William et al. \citep{williams1989learning} proposed an algorithm named Teacher forcing in their paper on training recurrent neural networks back in 1989. This approach suffers from over-generalization on training data and performs worse during inference. Later in 2015, Bengio et al. \citep{bengio2015scheduled} proposed a new training mechanism in their paper, “scheduled sampling for sequence prediction”, also known as Curriculum Learning. 

Taking inspiration from the natural language processing field, we demonstrate the effectiveness of these methods in the field of numerical simulations. The most common way to train transient models is to roll out the whole trajectory using its own predictions right from the beginning of the training process and calculating loss over all the predictions \cite{li2020fourier}. In this study, we show that Teacher Forcing and Curriculum Learning techniques yield better results as compared to the regular training procedure. Specifically, we show that Curriculum Learning outperforms all other approaches during inference. 


\section{Method}
During training a model for any transient (time-series) problem, we predict a sequence of outputs and calculate the loss between predictions and ground truth. Let’s consider an example, shown in Eq. \ref{eq1}, of a general transient system, where we can map solutions from $t_1, t_2, …, t_n$ to $t_{n+1}$ using a neural network. 
\begin{equation}
    \label{eq1}
    t'_{n+1} = \Theta \left( t_n, t_{n-1}, t_{n-2}, …, t_1 \right)
\end{equation}

Here $\Theta$ represents the parameters of the model. To predict the next time step, $t_{n+2}$, we pass in the previous $n$ steps consisting of its own prediction. In Eq. \ref{eq2}, the solution prediction of $t_{n+2}$ inputs the latest prediction at $t’_{n+1}$ as well as the remaining n-1 historical solutions. This approach is also used by Zongyi Li et al., in their work on Fourier Neural Operators \cite{li2020fourier}. 
\begin{equation}
    \label{eq2}
    t'_{n+2} = \Theta \left( t'_{n+1}, t_n, t_{n-1}, t_{n-2}, …, t_2 \right)
\end{equation}
During training time, as the ground truth for $t_{n+1}$ is available, we can use the ground truth solution, as shown in Eq. \ref{eq3}. As a result, the model always gets the correct sequence of previous $n$ time steps during training. This approach of handling time history is known as Teacher Forcing \cite{williams1989learning}. 
\begin{equation}
    \label{eq3}
    t'_{n+2} = \Theta \left( t_{n+1}, t_n, t_{n-1}, t_{n-2}, …, t_2 \right)
\end{equation}
The models trained with this approach often perform better than the approach used in Eq. \ref{eq2}. This is because the training in the previous approach is less robust due to noisy loss formulations resulting from error accumulation over long-time ranges. However, a model trained with teacher forcing never learns to correct itself as it never sees its own prediction during training. This results in divergence from the correct behavior for long-time predictions. 

\begin{figure}[h!]
 \centering
  \includegraphics[width=\textwidth]{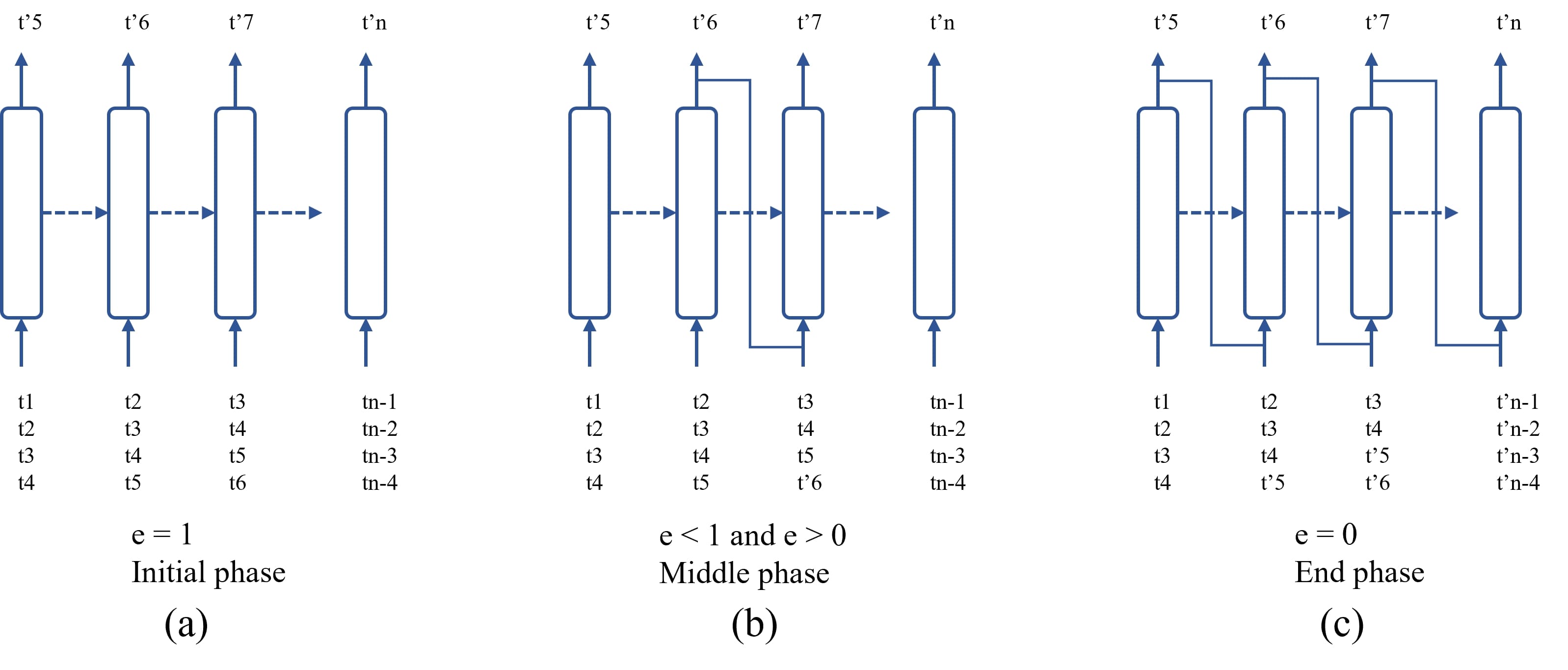}
  \caption{The model training starts with e=1, where predictions are made based on ground truth (a). In the middle phase, (b), the model sometimes takes ground truth and sometimes take its own prediction. This phase helps model to correct itself based on its own prediction. In the end, (c), the model is trained only with its own prediction. The horizontal dotted lines indicate that the same model weights are used for the next prediction.}
  \label{fig1}
\end{figure}

Curriculum Learning \cite{bengio2015scheduled}, an advanced variation of Teacher Forcing algorithm, is a mix of the approaches stated in the Eqs. \ref{eq2} and \ref{eq3}. In this approach, we randomly decide whether to use $t_{n+1}$ (target) or $t’_{n+1}$ (prediction) for future predictions during each training epoch. Fig. \ref{fig1} explains the curriculum learning approach in more detail. Let e be the ratio of number of targets used to the length of the sequence. In the initial phase of the training (when e=1), there is a heavier emphasis on using the target as shown in the Fig. \ref{fig1}(a), where all future predictions are computed from historical ground truth solutions. In the middle phase of the training, e begins to decrease and the number of predicted solutions used for future predictions are more. For example, in Fig. \ref{fig1}b the prediction of $t'_7$ depends on the previous prediction $t'_6$. Lower e values corresponds to use of more predicted solutions for future predictions. Finally, as shown in Fig. \ref{fig1}c, during the end phase of training all future predictions are computed from previously predicted solutions. The slow transition from using targets to predictions in future predictions allows for stable training and accurate models. Moreoever, the training imitates the actual inference process and hence, improves the robustness of the model

\section{Experiments}
In this work, we use a publicly available dataset to demonstrate the effectiveness of Curriculum Learning technique. The dataset solves the 2D vorticity equation derived from Navier-Stokes for a viscous, incompressible fluid on a unit torus. The viscosity is 1e-3 and the initial condition is solved for 50 timesteps on a 64 x 64 structured grid. More information about the dataset can be found in the work by Zyongi et al. \cite{li2020fourier}. The dataset has 5000 samples. 4000 samples are used for training, 500 samples are used for validation and 500 samples are used for testing. Out of 50 timesteps, only the first 40 are used for training. The last 10 timesteps are used to check the extrapolation performance.

There are two models used, a UNet model \cite{ronneberger2015u} and FNO 2D time model \cite{li2020fourier}. These models are trained with three different schemes. 
\begin{itemize}
    \item The whole roll-out using model predictions (This is similar to the training mechanics used in \cite{li2020fourier}) 
    \item Teacher forcing 
    \item Curriculum Learning 
\end{itemize}

During training, the model takes the first 10 timesteps as input and predicts the next one timestep. The model is rolled out for all the remaining 30 time steps before the gradient descent step. At the time of inference, the model is rolled out on all the 40 timesteps in the same fashion. The mean squared error loss is calculated on the whole sequence of 30 timesteps predictions. The model is trained for 500 epochs with Adam optimizer. The initial learning rate is 0.001 and is subsequently halved after every 100 epochs. For Fourier Neural Operator model, the hyperparameters used are same as mentioned \cite{li2020fourier}. All experiments are run on a single Nvidia Tesla A100 GPU. The decay scheme used for e, in Curriculum Learning, is linear.  

\section{Results}
Table \ref{tab1} shows the relative L2 norm on the test set. The numbers in the table are averaged over 500 test samples. The first technique gives satisfactory results. The Teacher forcing models show better results. The FNO model improves by 32.5\%. On the other hand, UNet shows a remarkable improvement of 50\%. 

\begin{table}[h]
\caption{Relative L2 norm comparison for different solution approaches applied to UNet and FNO}
\centering
\begin{tabular}{ |p{8cm}|p{2cm}|p{2cm}|  }
\hline
Solution approaches & FNO 2D time & UNet\\
\hline
Roll-out using model predictions  & 0.046 & 0.082\\
Teacher Forcing   & 0.031   & 0.041 \\
Curriculum Learning &   0.025 & 0.027 \\
\hline
\end{tabular}
\label{tab1}
\end{table} 

The Teacher Forcing improves the performance across all models. Curriculum Learning further improves the error rate. Curriculum Learning outperforms all other approaches. The FNO model improves by 52\% over the baseline. It achieves the best performance. UNet also improves by 67\% over its baseline. It may be observed that the UNet model is benefitted the most. With general training mechanics, the FNO model is inherently better in mapping transient problems than UNet, but with curriculum learning the results from UNet and FNO are very close. 

\begin{figure}[h!]
 \centering
  \includegraphics[width=\textwidth]{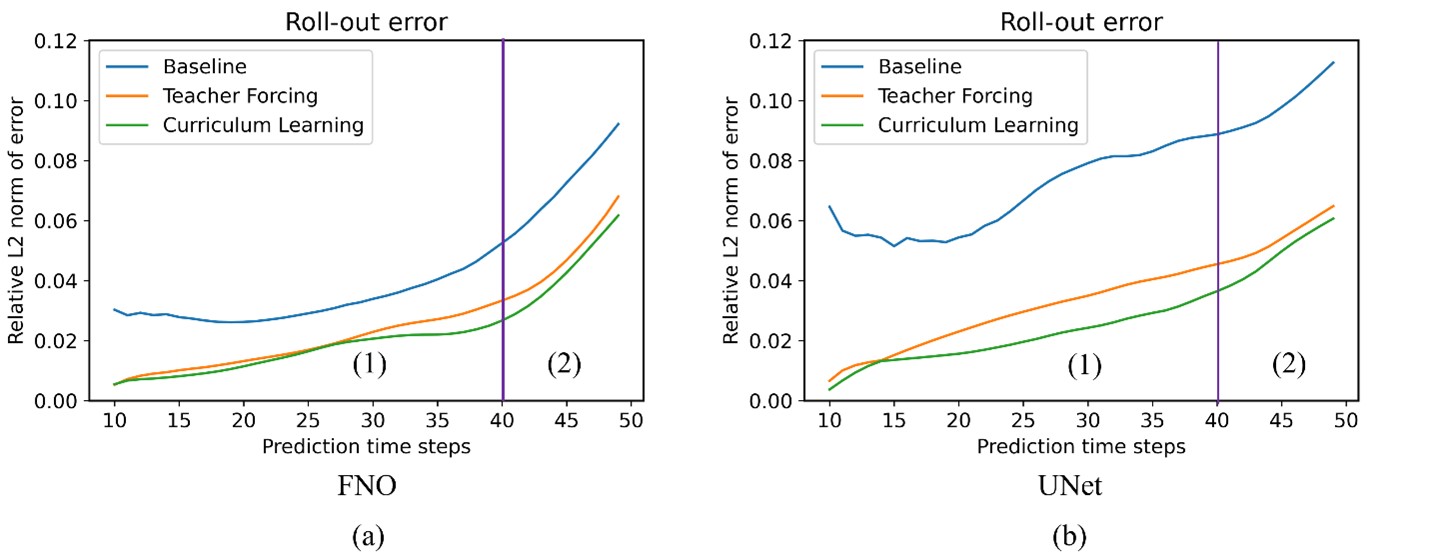}
  \caption{Roll-out error. The graphs (a) and (b) shows the average error over 500 test samples for FNO and UNet model respectively. The vertical line at time step number 40 represents the separation between interpolation and extrapolation region. (1) shows interpolation region whereas (2) shows extrapolation region.}
  \label{fig2}
\end{figure}

The following graphs in Fig. \ref{fig2} show the roll-out error on the test set. It is evident that the roll-out error improves with Curriculum Learning. The error seems to rise with a significantly smaller rate compared to corresponding baselines. Also, in the extrapolation region (time steps beyond 40), the model tends to behave reasonably well. As the data is unseen for the model, the error rate increases in this region for both models

\begin{figure}[h!]
 \centering
  \includegraphics[width=\textwidth]{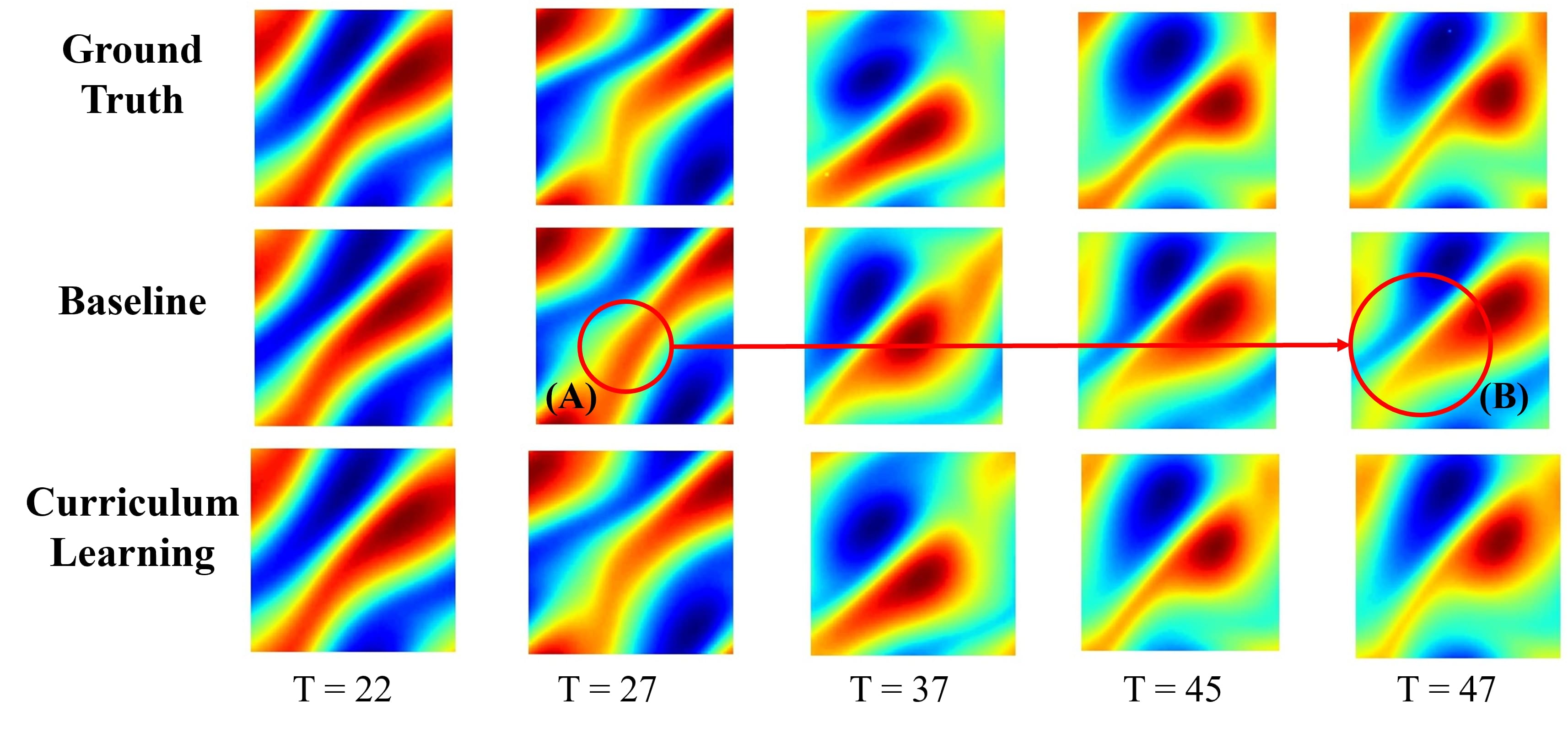}
  \caption{Error accumulation visualization. The figure shows the evolution of the solution for the baseline and Curriculum learning.}
  \label{fig3}
\end{figure}

In Fig. \ref{fig3}, we show some of the snapshots of the output produced by the UNet model for a test sample. The figure shows targets, baseline, and curriculum learning outputs for certain time steps. In the figure, circle (A) represents the approximate location in time and space where the prediction starts to deviate from the ground truth. This deviation accumulates and diverges from the actual trajectory as evident in circle (B). The curriculum learning output also has some deviation in the beginning, but it sufficiently corrects itself and does not deviate vastly from the target. One thing to note here is that the last 2 time steps are from the extrapolation region which the model never learned during the training. This indicates that the model’s performance significantly improves in the extrapolation region using this training mechanism.

\section{Conclusion and Future work} \label{conclusion}
This work shows the effectiveness of Curriculum Learning in learning PDEs for better performance during inference. It also shows that the techniques from other fields like Natural Language Processing can be beneficial in deep learning for numerical simulations. The simple change in the training mechanics help in better generalization and extrapolation as the model gradually learns to correct itself during training using its own predictions. This is possible because the training objective slowly gets similar to the inference. We expect that this method can be used for any generalized transient problems to improve the overall performance for any model architecture. Future work includes trying out different decay schemes like exponential or inverse sigmoid for curriculum learning. The investigation of the effect of number of epochs on the convergence could be useful in case of challenging datasets. The relationship between the learning rate decay scheme and the performance improvement with this method needs to be studies for stable convergence. 

\bibliographystyle{unsrt}
\bibliography{references}

\end{document}